# Semi-Dense 3D Reconstruction with a Stereo Event Camera

Yi Zhou[1,2], Guillermo Gallego[3], Henri Rebecq[3]
Laurent Kneip[4], Hongdong Li[1,2], Davide Scaramuzza[3]

[1]Australian National University
[2]Australian Centre for Robotic Vision
[3]Dept. Informatics and Dept. Neuroinformatics, University of Zurich and ETH Zurich
[4]School of Information Science and Technology, ShanghaiTech University

**Abstract.** Event cameras are bio-inspired sensors that offer several advantages, such as low latency, high-speed and high dynamic range, to tackle challenging scenarios in computer vision. This paper presents a solution to the problem of 3D reconstruction from data captured by a stereo event-camera rig moving in a static scene, such as in the context of stereo Simultaneous Localization and Mapping. The proposed method consists of the optimization of an energy function designed to exploit small-baseline spatio-temporal consistency of events triggered across both stereo image planes. To improve the density of the reconstruction and to reduce the uncertainty of the estimation, a probabilistic depth-fusion strategy is also developed. The resulting method has no special requirements on either the motion of the stereo event-camera rig or on prior knowledge about the scene. Experiments demonstrate our method can deal with both texture-rich scenes as well as sparse scenes, outperforming state-of-the-art stereo methods based on event data image representations.

## Multimedia Material

A supplemental video for this work is available at `https://youtu.be/Qrnpj2FD1e4`

## 1 Introduction

Event cameras, such as the Dynamic Vision Sensor (DVS) [1], are novel devices that output pixel-wise intensity changes (called "events") asynchronously, at the time they occur. As opposed to standard cameras, they do not acquire entire image frames, nor do they operate at a fixed frame rate. This asynchronous and differential principle of operation reduces power and bandwidth requirements drastically. Endowed with microsecond temporal resolution, event cameras are able to capture high-speed motions, which would typically cause severe motion blur on standard cameras. In addition, event cameras have a very High Dynamic Range (HDR) (e.g., 140 dB compared to 60 dB of most standard cameras), which allows them to be used on a broad illumination range. Hence, event cameras open the door to tackle challenging scenarios that are inaccessible to standard cameras, such as high-speed and/or HDR tracking [2–8], control [9,10] and Simultaneous Localization and Mapping (SLAM) [11–16].

Because existing computer vision algorithms designed for standard cameras do not directly apply to event cameras, the main challenge in visual processing with these novel sensors is to devise specialized algorithms that can exploit the temporally asynchronous and spatially sparse nature of the data produced by event cameras to unlock their potential. Some preliminary works addressed this issue by combining event cameras with additional sensors, such as standard cameras [8, 17, 18] or depth sensors [17, 19], to simplify the estimation task at hand. Although this approach obtained certain success, the capabilities of event cameras were not fully exploited since parts of such combined systems were limited by the lower dynamic range or slower devices. In this work, we tackle the problem of stereo 3D reconstruction for visual odometry (VO) or SLAM using event cameras alone. Our goal is to unlock the potential of event cameras by developing a method based on their working principle and using only events.

### 1.1   Related work on Event-based Depth Estimation

The majority of works on depth estimation with event cameras target the problem of "instantaneous" stereo, i.e., 3D reconstruction using events from a pair of synchronized cameras in stereo configuration (i.e., with a fixed baseline), during a very short time (ideally, on a per-event basis). Some of these works [20–22] follow the classical paradigm of solving stereo in two steps: epipolar matching followed by 3D point triangulation. Temporal coherence (e.g., simultaneity) of events across both left and right cameras is used to find matching events, and then standard triangulation [23] recovers depth. Other works, such as [24, 25], extend cooperative stereo [26] to the case of event cameras. These methods are typically demonstrated in scenes with static cameras and few moving objects, so that event matches are easy to find due to uncluttered event data.

Some works [27, 28] also target the problem of instantaneous stereo (depth maps produced using events over very short time intervals), but they use two non-simultaneous event cameras. These methods exploit a constrained hardware setup (two rotating event cameras with known motion) to either (*i*) recover intensity images on which conventional stereo is applied [27] or (*ii*) match events across cameras using temporal metrics and then use triangulation [28].

Recently, depth estimation with a *single* event camera has been shown in [11–14, 29]. These methods recover a semi-dense 3D reconstruction of the scene by integrating information from the events of a moving camera over a longer time interval, and therefore, require information of the relative pose between the camera and the scene. Hence, these methods do not target the problem of instantaneous depth estimation but rather the problem of depth estimation for VO or SLAM.

***Contribution.*** This paper is, to the authors' best knowledge, the first one to address the problem of non-instantaneous 3D reconstruction with a pair of event cameras in stereo configuration. Our approach is based on temporal coherence of events across left and right image planes. However, it differs from previous efforts (such as the instantaneous stereo methods [20–22, 27, 28]) in that: (*i*) we do not follow the classical paradigm of event matching plus triangulation, but rather a forward-projection approach that allows us to estimate depth without explicitly solving the event matching problem, (*ii*) we are able to handle sparse scenes (events generated by few objects) as well as cluttered scenes (events constantly generated everywhere in the image plane due to the motion of the camera), and (*iii*) we use camera pose information to integrate observations over time to produce semi-dense depth maps. Moreover, our method computes continuous depth values, as opposed to other methods, such as [11], which discretize the depth range.

***Outline***. Section 2 presents the 3D reconstruction problem considered and our solution, formulated as the minimization of an objective function that measures the temporal inconsistency of event time-surface maps across left and right image planes. Section 3 presents an approach to fuse multiple event-based 3D reconstructions into a single depth map. Section 4 evaluates our method on both synthetic and real event data, showing its good performance. Finally, Section 5 concludes the paper.

## 2  3D Reconstruction by Event Time-Surface Maps Energy Minimization

Our method is inspired by multi-view stereo pipelines for conventional cameras, such as DTAM [30], which aim at maximizing the photometric consistency through a number of narrow-baseline video frames. However, since event cameras do not output absolute intensity but rather intensity changes (the "events"), the direct photometric-consistency-based method cannot be readily applied. Instead, we exploit the fact that event cameras encode visual information in the form of microsecond-resolution timestamps of intensity changes.

For a stereo event camera, a detectable[1] 3D point in the overlapping field of view (FOV) of the cameras will generate an event on both left and right cameras. Ideally, these two events should spike simultaneously and their coordinates should be corresponding in terms of the epipolar geometry defined by both cameras. This property actually enables us to apply (and modify) an idea similar to DTAM, simply by replacing the photometric consistency with the stereo temporal consistency. However, as shown in [31], stereo temporal consistency does not strictly hold at the pixel level because of signal latency and jitter effects. Hence, we define our stereo temporal consistency criterion by aggregating measurements over spatio-temporal neighborhoods, rather than by comparing the event timestamps at two individual pixels, as we show next.

---

[1] A point at an intensity edge (i.e., non-homogeneous region of space), so that intensity changes (i.e., events) are generated when the point moves relative to the camera.

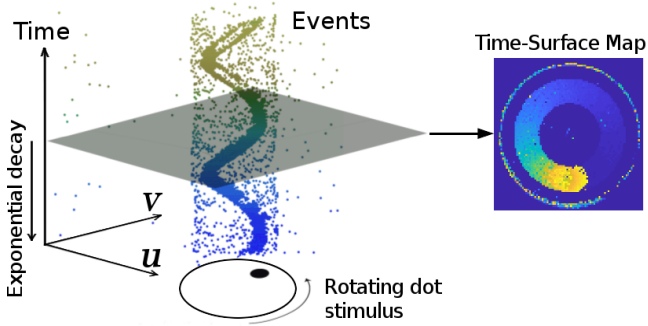

Fig. 1: Left: output of an event camera when viewing a rotating dot. Right: Time-surface map (1) at a time $t$, $\mathcal{T}(\mathbf{x}, t)$, which essentially measures how far in time (with respect to $t$) the last event spiked at each pixel $\mathbf{x} = (u, v)^T$. The brighter the color, the more recently the event was generated. Figure adapted from [33].

### 2.1   Event Time-Surface Maps

We propose to apply patch-match to compare a pair of spike-history maps, in place of the photometric warping error as used in DTAM [30]. Specifically, to create two distinctive maps, we advocate the use of *Time-Surface* inspired by [32] for event-based pattern recognition. As illustrated in Fig. 1, the output of an event camera is a stream of events, where each event $e_k = (u_k, v_k, t_k, p_k)$ consists of the space-time coordinates where the intensity change of predefined size happened and the sign (polarity $p_k \in \{+1, -1\}$) of the change[2]. The time-surface map at time $t$ is defined by applying an exponential decay kernel on the last spiking time $t_{\text{last}}$ at each pixel coordinate $\mathbf{x} = (u, v)^T$:

$$\mathcal{T}(\mathbf{x}, t) \doteq \exp\left(-\frac{t - t_{\text{last}}(\mathbf{x})}{\delta}\right), \tag{1}$$

where $\delta$, the decay rate parameter, is a small constant number (e.g., 30 ms in our experiments). For convenient visualization and processing, (1) is further rescaled to the range $[0, 255]$. Our objective function is constructed on a set of time-surface maps (1) at different observation times $t = \{t_s\}$.

### 2.2   Problem Formulation

We follow a global energy minimization framework to estimate the inverse depth map $\mathcal{D}$ in the reference view (RV) from a number of stereo observations $s \in \mathcal{S}_{\text{RV}}$ nearby. A *stereo observation* at time $t$ refers to a pair of time-surface maps created using (1), $(\mathcal{T}_{\text{left}}(\cdot, t), \mathcal{T}_{\text{right}}(\cdot, t))$. A stereo observation could be triggered by either a pose update or at a constant rate. For each pixel $\mathbf{x}$ in the reference view, its inverse depth $\rho^\star \doteq 1/z^\star$ is estimated by optimizing the objective function:

$$\rho^\star = \arg\min_\rho C(\mathbf{x}, \rho) \tag{2}$$

---

[2] Event polarity is not used, as [13] shows that it is not needed for 3D reconstruction.

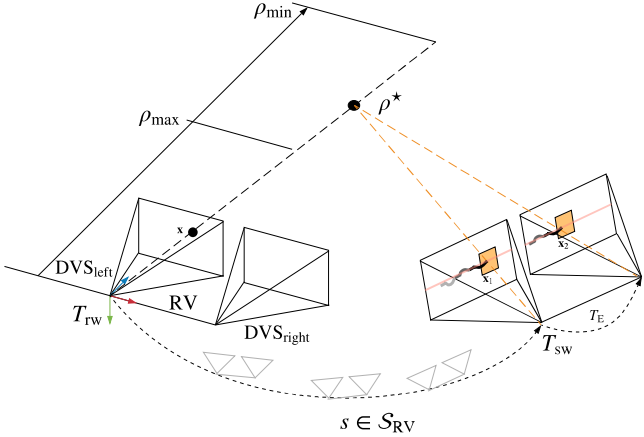

Fig. 2: Illustration of the geometry of the proposed problem and solution. The reference view (RV) is on the left, in which an event with coordinates $\mathbf{x}$ is back-projected into 3D space with a hypothetical inverse depth $\rho$. The optimal inverse depth $\rho^\star$, lying inside the search interval $[\rho_{\min}, \rho_{\max}]$, corresponds to the real location of the 3D point which fulfills the temporal consistency in each neighboring stereo observation $s$.

$$C(\mathbf{x}, \rho) \doteq \frac{1}{|\mathcal{S}_{\mathrm{RV}}|} \sum_{s \in \mathcal{S}_{\mathrm{RV}}} \|\tau_{\mathrm{left}}^s(\mathbf{x}_1(\rho)) - \tau_{\mathrm{right}}^s(\mathbf{x}_2(\rho))\|_2^2, \qquad (3)$$

where $|\mathcal{S}_{\mathrm{RV}}|$ denotes the number of involved neighboring stereo observations used for averaging. The function $\tau_{\mathrm{left/right}}^s(\mathbf{x})$ returns the temporal information $\mathcal{T}_{\mathrm{left/right}}(\cdot, t)$ inside a $w \times w$ patch centered at image point $\mathbf{x}$. The residual

$$r_s(\rho) \doteq \|\tau_{\mathrm{left}}^s(\mathbf{x}_1(\rho)) - \tau_{\mathrm{right}}^s(\mathbf{x}_2(\rho))\|_2 \qquad (4)$$

denotes the temporal difference in $l_2$ norm between patches centered at $\mathbf{x}_1$ and $\mathbf{x}_2$ in the left and right event cameras, respectively.

The geometry behind the proposed objective function is illustrated in Fig. 2. Since we assume the calibration (intrinsic and extrinsic parameters) as well as the pose of the left event camera $T_{sr}$ at each observation are known (e.g., from a tracking algorithm such as [12,14]), the points $\mathbf{x}_1$ and $\mathbf{x}_2$ are given by $\mathbf{x}_1(\rho) = \pi(T_{sr}\pi^{-1}(\mathbf{x}, \rho))$ and $\mathbf{x}_2(\rho) = \pi(T_{\mathrm{E}}T_{sr}\pi^{-1}(\mathbf{x}, \rho))$, respectively. The function $\pi : \mathbb{R}^3 \to \mathbb{R}^2$ projects a 3D point onto the camera's image plane, while its inverse function $\pi^{-1} : \mathbb{R}^2 \to \mathbb{R}^3$ back-projects a pixel into 3D space given the inverse depth $\rho$. $T_{\mathrm{E}}$ denotes the transformation from the left to the right event camera. Note that all event coordinates $\mathbf{x}$ are undistorted and rectified.

To verify that the proposed objective function (3) does lead to the optimum depth for a generic event in the reference view (Fig. 3(a)), a number of stereo observations from a real stereo event-camera sequence [34] have been created (Figs. 3(c) and 3(d)) and used to visualize the energy at the event location (Fig. 3(b)). The size of the patch is $w = 25$ pixels throughout the paper.

6      Y. Zhou, G. Gallego, H. Rebecq, L. Kneip, H. Li, D. Scaramuzza

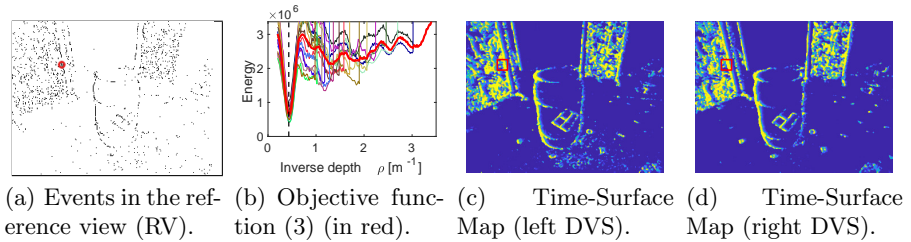

(a) Events in the reference view (RV).   (b) Objective function (3) (in red).   (c) Time-Surface Map (left DVS).   (d) Time-Surface Map (right DVS).

Fig. 3: Proposed objective function. (a) A randomly selected event, at pixel $\mathbf{x}$, is marked by a red circle in the reference view. The energy $C(\mathbf{x}, \rho)$ in (3) is visualized in (b) as a function of $\rho$, with the thick red curve obtained by averaging the costs $C(\mathbf{x}_i, \rho)$ of neighboring pixels $\mathbf{x}_i$ in a patch centered at $\mathbf{x}$ (indicated by curves with random colors). The vertical dashed line (black) indicates the ground truth inverse depth. The time-surface maps of the left and the right event cameras at one of the observation times are shown in (c) and (d), respectively, where the patches for measuring the temporal residual are indicated in red.

Note that our approach significantly departs from classical two-step event-processing methods [20–22] that solve the stereo matching problem first and then triangulate the 3D point, which is prone to errors due to the difficulty in establishing correct event matches during very short time intervals. These two-step approaches work in a "back-projection" fashion, mapping 2D event measurements to 3D space. Instead, our approach combines matching and triangulation in a single step, operating in a forward-projection manner (from 3D space to 2D event measurements). As shown in Fig. 2, an inverse depth hypothesis $\rho$ yields a 3D point, $\pi^{-1}(\mathbf{x}, \rho)$, whose projection on both stereo image planes for all times "$s$" gives curves $\mathbf{x}_1^s(\rho)$ and $\mathbf{x}_2^s(\rho)$ that are compared in the objective function (3). Hence, an inverse depth hypothesis $\rho$ establishes candidate stereo event matches, and the best matches are obtained once the objective function has been minimized with respect to $\rho$.

### 2.3  Inverse Depth Estimation

The proposed objective function (3) is optimized using non-linear least squares methods. The Gauss-Newton method is used here, which iteratively discovers the root of the necessary optimality condition

$$\frac{\partial C}{\partial \rho} = \frac{2}{|\mathcal{S}_{\text{RV}}|} \sum_{s \in \mathcal{S}_{\text{RV}}} r_s \frac{\partial r_s}{\partial \rho} = 0. \tag{5}$$

Substituting the linearization of $r_s$ at $\rho_k$ using the first order Taylor formula, $r_s(\rho_k + \Delta\rho) \approx r_s(\rho_k) + J_s(\rho_k)\Delta\rho$, in (5) we obtain

$$\sum_{s \in \mathcal{S}_{\text{RV}}} J_s(r_s + J_s \Delta\rho) = 0, \tag{6}$$

where both, residual $r_s \equiv r_s(\rho_k)$ and Jacobian $J_s \equiv J_s(\rho_k)$, are scalars. Consequently the inverse depth $\rho$ is iteratively updated by adding the increment

$$\Delta\rho = -\frac{\sum_{s \in \mathcal{S}_{\text{RV}}} J_s r_s}{\sum_{s \in \mathcal{S}_{\text{RV}}} J_s^2}. \tag{7}$$

The Jacobian is computed by applying the chain rule,

$$\begin{aligned} J_s(\rho) &\doteq \frac{\partial}{\partial \rho} \|\mathcal{T}_{\text{left}}^s(\mathbf{x}_1(\rho)) - \mathcal{T}_{\text{right}}^s(\mathbf{x}_2(\rho))\|_2 \\ &= \frac{1}{\|\mathcal{T}_{\text{left}}^s - \mathcal{T}_{\text{right}}^s\|_2 + \epsilon} \left(\mathcal{T}_{\text{left}}^s - \mathcal{T}_{\text{right}}^s\right)_{1 \times w^2}^T \left(\frac{\partial \mathcal{T}_{\text{left}}^s}{\partial \rho} - \frac{\partial \mathcal{T}_{\text{right}}^s}{\partial \rho}\right)_{w^2 \times 1}, \end{aligned} \tag{8}$$

where, for simplicity, the pixel notation $\mathbf{x}_i(\rho)$ is omitted in the last equation. To avoid division by zero, a small number $\epsilon$ is added to the length of the residual vector. Actually, as shown by an investigation on the distribution of the temporal residual $r_s$ in Section 3.1, the temporal residual is unlikely to be close to zero for valid stereo observations (i.e., patches with enough events). The derivative of the time-surface map with respect to the inverse depth is calculated by

$$\frac{\partial \boldsymbol{\tau}^s}{\partial \rho} = \frac{\partial \boldsymbol{\tau}^s}{\partial \mathbf{x}} \frac{\partial \mathbf{x}}{\partial \rho} = \left(\frac{\partial \boldsymbol{\tau}^s}{\partial u}, \frac{\partial \boldsymbol{\tau}^s}{\partial v}\right)_{w^2 \times 2} \left(\frac{\partial u}{\partial \rho}, \frac{\partial v}{\partial \rho}\right)^T. \tag{9}$$

The computation of $\partial u/\partial \rho$ and $\partial v/\partial \rho$ is given in the supplementary material.

The overall procedure is summarized in Algorithm 1. The inputs of the algorithm are, respectively, the pixel coordinate $\mathbf{x}$ of an event in the RV, a set of stereo observations (time-surface maps) $\mathcal{T}_{\text{left/right}}^s$ ($s \in \mathcal{S}_{\text{RV}}$), the relative pose $T_{sr}$ from the RV to each involved stereo observation $s$ and the constant extrinsic parameters between both event cameras, $T_{\text{E}}$. The inverse depths of all events in the RV are estimated independently. Therefore, the computation is parallelizable. The basin of convergence is first localized by a coarse search over the range of plausible inverse depth values followed by a nonlinear refinement using the Gauss-Newton method. The coarse search step is selected to balance efficiency and accuracy when locating the basin of convergence, and is also based on our observation that the width of the basin is always bigger than $0.2\,\text{m}^{-1}$ for the experiments carried out.

## 3   Semi-Dense Reconstruction

The 3D reconstruction method presented in Section 2 produces a sparse depth map at the reference view (RV). To improve the density of the reconstruction while reducing the uncertainty of the estimated depth, we run the reconstruction method (Algorithm 1) on several RVs along time and fuse the results. To this end, the uncertainty of the inverse depth estimation is studied in this section. Based on the derived uncertainty, a fusion strategy is developed and is incrementally applied as sparse reconstructions of new RVs are obtained. Our final reconstruction approaches a semi-dense level as it reconstructs depth for all pixels that lie along edges.

**Algorithm 1** Inverse Depth Estimation at a Reference View (RV)
─────────────────────────────────────────────────────
1: **Input**: pixel **x**, stereo event observations $\mathcal{T}^s_{\text{left}}, \mathcal{T}^s_{\text{right}}$ and poses $T_{sr}, T_{\text{E}}$.
2: $\rho_0 \leftarrow \rho_{\text{initial}}$ (by coarse search over a range $[\rho_{\min}, \rho_{\max}]$).
3: **while** not converged **do**
4:     **for** each observation $s$ **do**
5:         Compute $r_s(\rho_k)$ in (4).
6:         Compute $J_s(\rho_k)$ using (8).
7:     **end for**
8:     Update: $\rho_k \leftarrow \rho_k + \Delta\rho$, using (7).
9: **end while**
10: **return** Inverse depth $\rho_k$.
─────────────────────────────────────────────────────

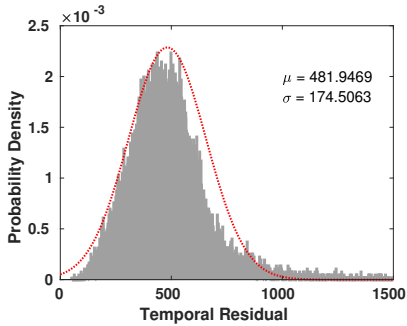

Fig. 4: Probability distribution (PDF) of the temporal residuals $\{r_i\}$: empirical (gray) and Gaussian fit $\mathcal{N}(\mu, \sigma^2)$ (red line).

### 3.1 Uncertainty of Inverse Depth Estimation

In the last iteration of Gauss-Newton's method, the inverse depth is updated by

$$\rho^\star \equiv \rho_k \leftarrow \rho_k + \Delta\rho(\mathbf{r}), \tag{10}$$

where $\Delta\rho$ is a function of the residuals $\mathbf{r} \doteq \{r_1, r_2, \ldots, r_s \,|\, s \in \mathcal{S}_{\text{RV}}\}$ as defined in (7). The variance $\sigma^2_{\rho^\star}$ of the inverse depth estimate can be derived using uncertainty propagation [35]. For simplicity, only the noise in the temporal residuals $\mathbf{r}$ is considered:

$$\sigma^2_{\rho^\star} \approx \left(\frac{\partial \rho^\star}{\partial \mathbf{r}}\right)^T (\sigma_r^2 \text{Id}) \frac{\partial \rho^\star}{\partial \mathbf{r}} = \frac{\sigma_r^2}{\sum_{s \in \mathcal{S}_{\text{RV}}} J_s^2}. \tag{11}$$

The derivation of this equation can be found in the supplementary material. We determine $\sigma_r$ empirically by investigating the distribution of the temporal residuals $\mathbf{r}$. Using the ground truth depth, we sample a large number of temporal residuals $\mathbf{r} = \{r_1, r_2, ..., r_n\}$. The variance $\sigma_r^2$ is obtained by fitting a Gaussian distribution to the histogram of $\mathbf{r}$, as illustrated in Fig. 4.

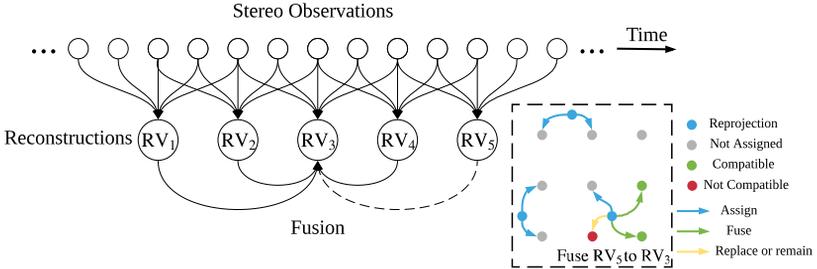

Fig. 5: Depth map fusion strategy. All stereo observations ($\mathcal{T}_{\text{left}}^s, \mathcal{T}_{\text{right}}^s$) are denoted by hollow circles and listed in chronological order. Neighboring RVs are fused into a chosen RV$^\star$ (e.g., RV$_3$). Using the fusion from RV$_5$ to RV$_3$ as an example, the fusion rules are illustrated in the dashed square, in which a part of the image plane is visualized. The blue dots are the reprojections of 3D points in RV$_5$ on the image plane of RV$_3$. Gray dots represent unassigned pixels which will be assigned by blue dots within one pixel away. Pixels that have been assigned, e.g., the green ones (compatible with the blue ones) will be fused. Pixels that are not compatible (in red) will either remain or be replaced, depending on which distribution has the smallest uncertainty.

### 3.2 Inverse Depth Fusion

To improve the density of the reconstruction, inverse depth estimates from multiple RVs are incrementally transferred to a selected reference view, RV$^\star$, and fused. Assuming the inverse depth of a pixel in RV$_i$ follows a distribution $\mathcal{N}(\rho_a, \sigma_a^2)$, its corresponding location in RV$^\star$ is typically a non-integer coordinate $\mathbf{x}^f$, which will have an effect on the four neighboring pixels coordinates $\{\mathbf{x}_j^i\}_{j=1}^4$. Using $\mathbf{x}_1^i$ as an example, the fusion is performed based on the following rules:

1. Assign $\mathcal{N}(\rho_a, \sigma_a^2)$ to $\mathbf{x}_1^i$ if no previous distribution exists.
2. If there is an existing inverse depth distribution assigned at $\mathbf{x}_1^i$, e.g., $\mathcal{N}(\rho_b, \sigma_b^2)$, the compatibility between the two inverse depth hypotheses is checked to decide whether they are fused. The compatibility is evaluated using the $\chi^2$ test at 95 % [35]:

$$\frac{(\rho_a - \rho_b)^2}{\sigma_a^2} + \frac{(\rho_a - \rho_b)^2}{\sigma_b^2} < 5.99. \qquad (12)$$

If the two hypotheses are compatible, they are fused into a single inverse depth distribution:

$$\mathcal{N}\left(\frac{\sigma_a^2 \rho_b + \sigma_b^2 \rho_a}{\sigma_a^2 + \sigma_b^2}, \frac{\sigma_a^2 \sigma_b^2}{\sigma_a^2 + \sigma_a^2}\right), \qquad (13)$$

otherwise the distribution with the smallest variance remains.

An illustration of the fusion strategy is given in Fig. 5.

## 4   Experiments

The proposed stereo 3D reconstruction method is evaluated in this section. We first introduce the configuration of our stereo event-camera system and the datasets used in the experiments. Afterwards, both quantitative and qualitative evaluations are presented. Additionally, the depth fusion process is illustrated to highlight how it improves the density of the reconstruction while reducing depth uncertainty.

### 4.1   Stereo Event-camera Setup

To evaluate our method, we use sequences from publicly available simulators [36] and datasets [34], and we also collect our own sequences using a stereo event-camera rig (Fig. 6). The stereo rig consists of two Dynamic and Active Pixel Vision Sensors (DAVIS) [37] of $240 \times 180$ pixel resolution, which are calibrated intrinsically and extrinsically[3] using *Kalibr* [38]. Since our algorithm is working on rectified and undistorted coordinates, the joint undistortion and rectification transformation are computed in advance.

As the stereo event-camera system moves, a new stereo observation $(\mathcal{T}^s_{\text{left}}, \mathcal{T}^s_{\text{right}})$ is generated when a pose update is available. The generation consists of two steps. The first step is to generate a rectified event map by collecting all events that occurred within 10 ms (from the pose's updating time to the past), as shown in Fig. 3(a). The second step is to refresh the time-surface maps in both left and right event cameras, as shown in Figs. 3(c) and 3(d). One of the observations is selected as the RV. The rectified event map of the RV together with the rest of the observations are fed to the inverse depth estimation module (Algorithm 1). We use the rectified event map as a selection map, i.e., we estimate depth values only at the pixels with non-zero values in the rectified event map (as shown in Figs. 6 (c) and (d)). As more and more RVs are reconstructed and fused together, the result becomes both more dense and more accurate.

### 4.2   Results

The evaluation is performed on six sequences, including a synthetic sequence from the simulator [36], three sequences collected by ourselves (hand-held) and two sequences from [34] (with a stereo event camera mounted on a drone). A snapshot of each scene is given in the first column of Fig. 7. In the synthetic sequence, the stereo event-camera system looks orthogonally towards three frontal parallel planes while performing a pure translation. Our three sequences showcase typical office scenes with various office supplies. The stereo event-camera rig is hand-held and performs arbitrary 6-DOF motion, which is recorded by a motion-capture system with sub-millimeter accuracy. The other two sequences

---
[3] The DAVIS comprises both a frame camera and an event sensor (DVS) aligned perfectly on the same pixel array. Hence, we calibrate the stereo pair using standard methods on the intensity frames.

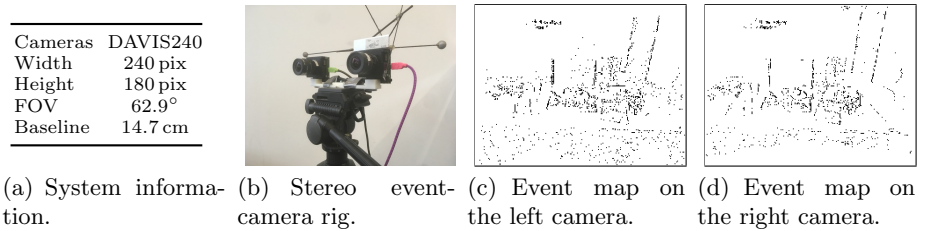

(a) System information.  (b) Stereo event-camera rig.  (c) Event map on the left camera.  (d) Event map on the right camera.

Fig. 6: Left, (a) and (b): the stereo event-camera rig used in our experiment, consisting of two synchronized DAVIS [37] devices. Right, (c) and (d): rectified event maps at one time observation.

Table 1: Quantitative evaluation on sequences with ground truth depth.

|  | Dataset | simulation_3planes [36] | Indoor_flying1 [34] | Indoor_flying3 [34] |
|---|---|---|---|---|
|  | Depth range | 2.76 m | 4.96 m | 5.74 m |
| Our Method | Mean error | **0.03** m | **0.13** m | **0.33** m |
|  | Median error | **0.01** m | **0.05** m | 0.11 m |
|  | Relative error | **1.17** % | **2.65** % | **5.79** % |
| FCVF [39] | Mean error | 0.05 m | 0.99 m | 1.03 m |
|  | Median error | 0.03 m | 0.25 m | **0.11** m |
|  | Relative error | 1.84 % | 20.8 % | 17.3 % |
| SGM [40] | Mean error | 0.08 m | 0.93 m | 1.19 m |
|  | Median error | 0.03 m | 0.31 m | 0.20 m |
|  | Relative error | 3.22 % | 18.7 % | 20.8 % |

are collected in a large indoor environment using a drone [34], with pose information also from a motion-capture system. These two sequences are very challenging for two reasons: (*i*) a wide variety of structures such as chairs, barrels, a tripod on a cabinet, etc. can be found in this scene, and (*ii*) the drone undergoes relatively high-speed motions during data collection.

Quantitative evaluation on datasets with ground truth depth are given in Table 1, where we compare our method with two state-of-the-art instantaneous stereo matching methods, "Fast Cost-Volume Filtering" (FCVF) [39] and "Semi-Global Matching" (SGM) [40], working on pairs of time-surface images (as in Figs. 3(c) and 3(d)). We report the *mean* depth error, the *median* depth error and the relative error (defined as the *mean* depth error divided by the depth range of the scene [13]). In fairness to the comparison, the fully dense depth maps returned by FCVF and SGM are masked by the non-zero pixels in the time-surface images. Besides, the boundary of the depth maps are cropped considering the block size used in each implementation. The best results per sequence are highlighted in bold in Table 1. Our method outperforms the other two competitors on all sequences. Although FCVF and SGM also give satisfactory results on the synthetic sequence, they do not work well in more complicated scenarios in which the observations are either not dense enough, or the temporal consistency does not strictly hold in a single stereo observation.

Reconstruction results on all sequences are visualized in Fig. 7. Images on the first column are raw intensity frames from the DAVIS. They convey the appearance of the scenes but are not used by our algorithm. The second column shows rectified and undistorted event maps in the left event-camera of a RV. The number of the events depends on not only on the motion of the stereo rig but also on the amount of visual contrast in the scene. Semi-dense depth maps (after fusion with several neighboring RVs) are given in the third column, pseudo-colored from red (close) to blue (far). The last column visualizes the 3D point cloud of each sequence at a chosen perspective. Note that only points whose variance $\sigma_\rho^2$ is smaller than $0.8 \times (\sigma_\rho^{\max})^2$ are visualized in 3D.

The reconstruction of the rectified events in one RV is sparse and, typically, full of noise. To show how the fusion strategy improves the density of the reconstruction as well as reduces the uncertainty, we additionally perform an experiment that visualizes the fusion process incrementally. As shown in Fig. 8, the first column visualizes the uncertainty maps before the fusion. The second to the fourth column demonstrates the uncertainty maps after fusing the result of a RV with its neighboring 4, 8 and 16 estimations, respectively. Hot colors refer to high uncertainty while cold colors mean low uncertainty. The result becomes increasingly dense and accurate as more and more RVs are fused. Note that the remaining highly uncertain estimates generally correspond to events that are caused by either noise or low-contrast patterns.

## 5   Conclusion

This paper has proposed a novel and effective solution to 3D reconstruction using a pair of temporally-synchronized event cameras in stereo configuration. This is, to the best of the authors' knowledge, the first one to address such a problem allowing stereo SLAM applications with event cameras. The proposed energy minimization method exploits spatio-temporal consistency of the events across cameras to achieve high accuracy (between 1% and 5% relative error), and it outperforms state-of-the-art stereo methods using the same spatio-temporal image representation of the event stream. Future work includes the development of a full stereo visual odometry system, by combining the proposed 3D reconstruction strategy with a stereo-camera pose tracker, in a parallel tracking and mapping fashion [14].

## Acknowledgment

The research leading to these results is supported by the Australian Centre for Robotic Vision and the National Center of Competence in Research (NCCR) Robotics, through the Swiss National Science Foundation, the SNSF-ERC Starting Grant and the NCCR Ph.D. Exchange Scholarship Programme. Yi Zhou also acknowledges the financial support from the China Scholarship Council for his Ph.D. Scholarship No. 201406020098.

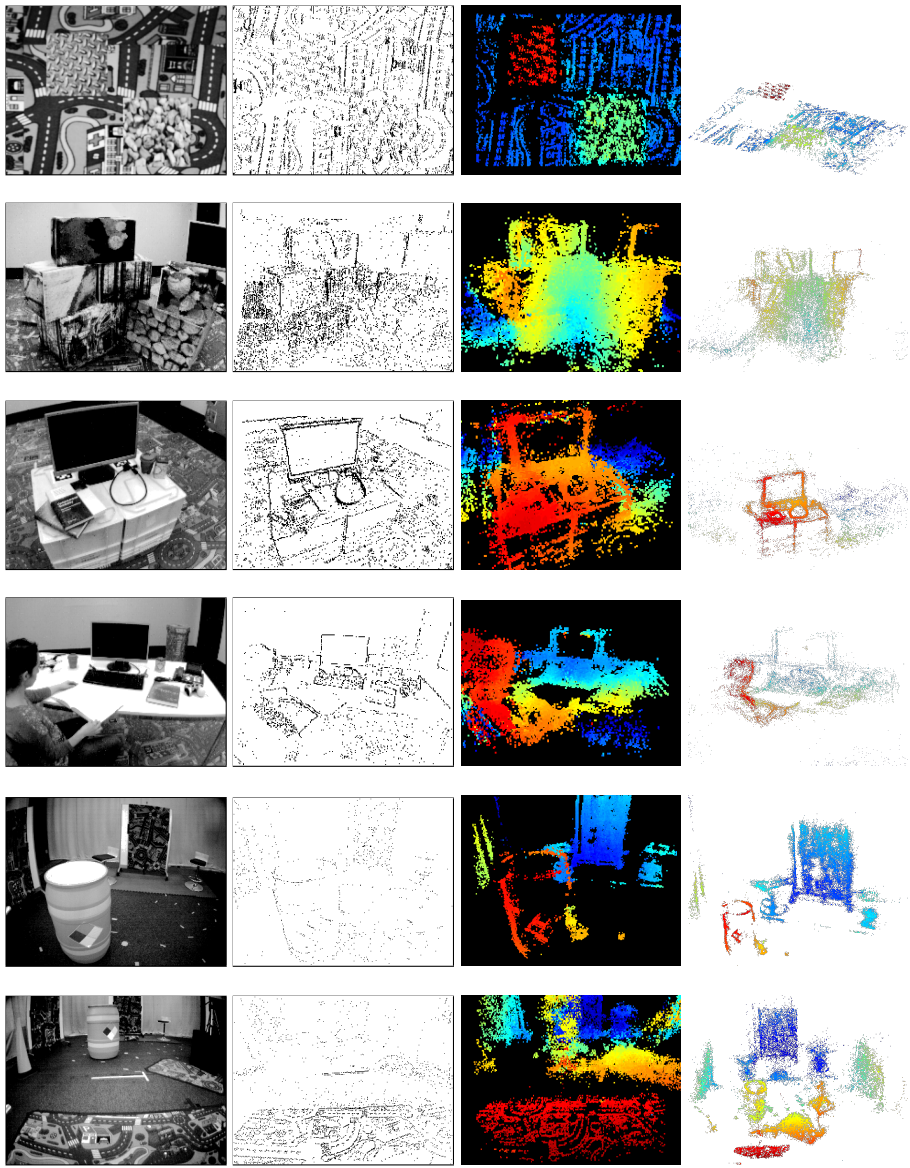

Fig. 7: Results of the proposed method on several datasets. Images on the first column are raw intensity frames (not rectified nor lens-distortion corrected). The second column shows the events (undistorted and rectified) in the left event camera of a reference view (RV). Semi-dense depth maps (after fusion with several neighboring RVs) are given in the third column, colored according to depth, from red (close) to blue (far). The fourth column visualizes the 3D point cloud of each sequence at a chosen perspective. No post-processing (such as regularization through median filtering [13]) was performed.

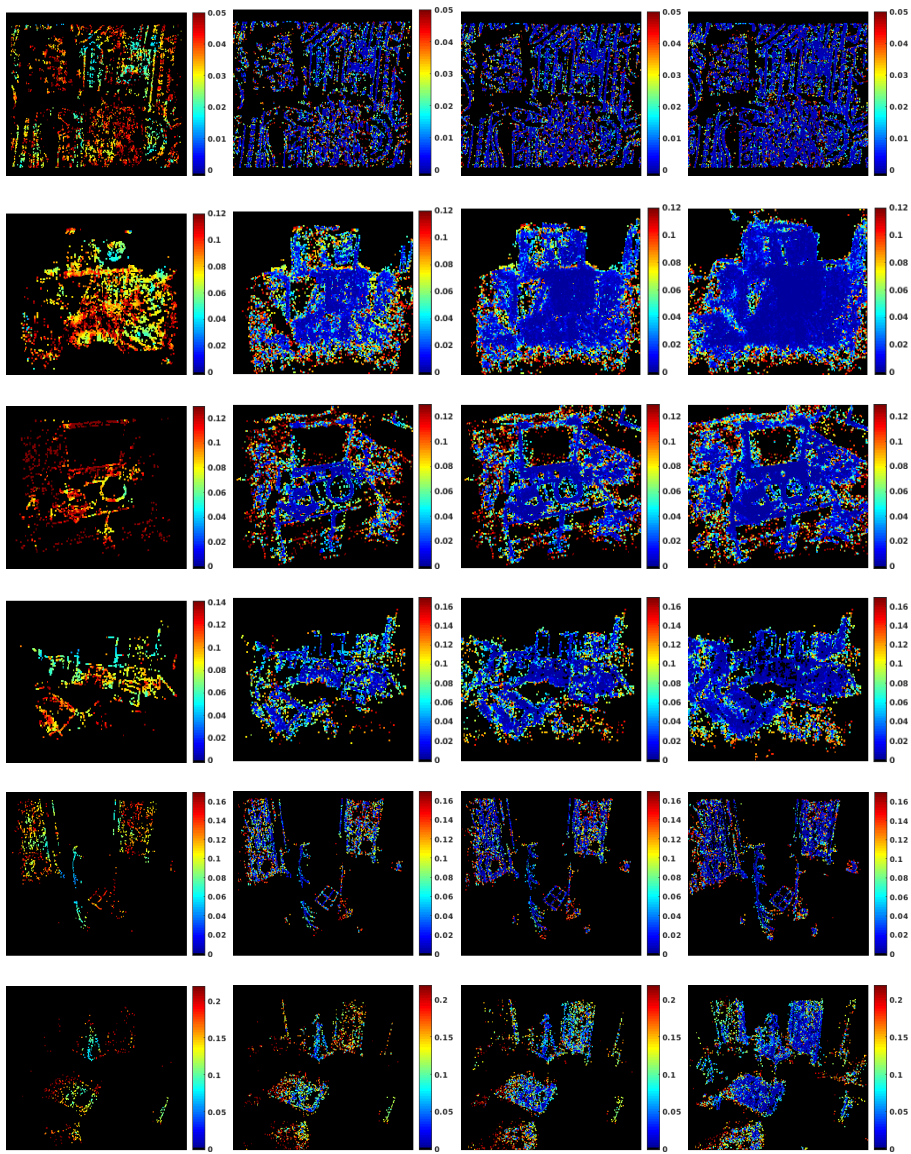

Fig. 8: Illustration of how the fusion strategy increasingly improves the density of the reconstruction while reducing depth uncertainty. The first column shows the uncertainty maps $\sigma_\rho$ before the fusion. The second to the fourth columns report the uncertainty maps after fusing with 4, 8 and 16 neighboring estimations, respectively.

Semi-Dense 3D Reconstruction with a Stereo Event Camera     1737. Brandli, C., Berner, R., Yang, M., Liu, S.C., Delbruck, T.: A 240x180 130dB 3us latency global shutter spatiotemporal vision sensor. IEEE J. Solid-State Circuits **49**(10) (2014) 2333–2341
38. Furgale, P., Rehder, J., Siegwart, R.: Unified temporal and spatial calibration for multi-sensor systems. In: IEEE/RSJ Int. Conf. Intell. Robot. Syst. (IROS). (2013)
39. Hosni, A., Rhemann, C., Bleyer, M., Rother, C., Gelautz, M.: Fast cost-volume filtering for visual correspondence and beyond. IEEE Trans. Pattern Anal. Machine Intell. **35**(2) (February 2013) 504–511
40. Hirschmuller, H.: Stereo processing by semiglobal matching and mutual information. IEEE Trans. Pattern Anal. Machine Intell. **30**(2) (February 2008) 328–341

## 6   Appendices (Supplementary Material)

### 6.1   Calculation of the Derivatives

The objective function requires to warp every event's location **x** in the reference view to each pair of involved stereo observation $\mathbf{x}_1$ and $\mathbf{x}_2$. First, the 3D point **p** inducing the event **x** is recovered by performing a back-projection, given the inverse depth $\rho$:

$$\dot{\mathbf{p}} = \frac{1}{\rho}\begin{pmatrix} \mathtt{P}_1 \\ 0\ 0\ 0\ z \end{pmatrix}^{-1} \begin{pmatrix} u \\ v \\ 1 \\ 1 \end{pmatrix} = \begin{pmatrix} \frac{u-p_{13}}{p_{11}\rho} \\ \frac{v-p_{23}}{p_{22}\rho} \\ \frac{1}{\rho} \\ 1 \end{pmatrix},$$

where $\mathtt{P}_1$ is the $3 \times 4$ projection matrix of the left event camera. The following calculation is based on the fact that the last column of $\mathtt{P}_1$ is $\mathbf{0}_{3\times 1}$. Transforming **p** to the left camera coordinate of an observation out of $\mathcal{S}_{\text{RV}}$ gives,

$$\mathbf{p}_1 = \mathtt{R}\mathbf{p} + \mathbf{t}. \tag{14}$$

The warping results are obtained by

$$\dot{\mathbf{x}}_1 = \mathtt{P}_1\mathbf{p}_1, \tag{15}$$

$$\dot{\mathbf{x}}_2 = \mathtt{P}_2\mathbf{p}_1. \tag{16}$$

Taking the left event camera, for example,

$$u_1 = \frac{A + B\rho}{C + D\rho}, \tag{17}$$

where

$$\begin{aligned} A &= (p_{11}r_{11} + p_{13}r_{31})\frac{u - p_{13}}{p_{11}} \\ &\quad + (p_{11}r_{12} + p_{13}r_{32})\frac{v - p_{23}}{p_{22}} + (p_{11}r_{13} + p_{13}r_{33}), \\ B &= p_{11}t_x + p_{13}t_z + p_{14}, \\ C &= \frac{r_{31}(u - p_{13})}{p_{11}} + \frac{r_{32}(v - p_{23})}{p_{22}} + r_{33}, \\ D &= t_z. \end{aligned} \tag{18}$$

Similarly,

$$v_1 = \frac{A' + B'\rho}{C' + D'\rho}, \tag{19}$$

with

$$\begin{aligned}
A' &= (p_{22}r_{21} + p_{23}r_{31})\frac{u - p_{13}}{p_{11}} \\
&\quad + (p_{22}r_{22} + p_{23}r_{32})\frac{v - p_{23}}{p_{22}} + (p_{22}r_{23} + p_{23}r_{33}), \\
B' &= p_{22}t_y + p_{23}t_z + p_{24}, \\
C' &= C \\
D' &= D.
\end{aligned} \quad (20)$$

Therefore, the derivatives with respect to inverse depth $d$ are:

$$\begin{aligned}
\frac{\partial u}{\partial \rho} &= \frac{BC - AD}{(C + D\rho)^2}, \\
\frac{\partial v}{\partial \rho} &= \frac{B'C' - A'D'}{(C' + D'\rho)^2}.
\end{aligned} \quad (21)$$

### 6.2 Computing Uncertainty Propagation, from the Event Residuals to the Estimated Inverse Depth

Following (10) and only considering the temporal residual for simplicity, we have

$$\rho^\star = \rho_k - \frac{1}{\gamma}\underbrace{(J_1 r_1 + J_2 r_2 + \cdots + J_s r_s)}_{s \in \mathcal{S}_{\text{RV}}}, \quad (22)$$

where

$$\gamma \doteq \sum_{s \in \mathcal{S}_{\text{RV}}} J_s^2. \quad (23)$$

Therefore the derivative of $\rho^\star$ with respect to $\mathbf{r}$ is

$$\frac{\partial \rho^\star}{\partial \mathbf{r}} = -\frac{1}{\gamma}(J_1, J_2, \cdots, J_s). \quad (24)$$

Substituting (24) in (11), the overall uncertainty of the inverse depth is, to first order, given by

$$\begin{aligned}
\sigma^2_{\rho^\star} &\approx \frac{1}{\gamma}(J_1, J_2, \cdots, J_s)\begin{pmatrix} \sigma_r^2 & & & \\ & \sigma_r^2 & & \\ & & \ddots & \\ & & & \sigma_r^2 \end{pmatrix}\frac{1}{\gamma}\begin{pmatrix} J_1 \\ J_2 \\ \vdots \\ J_s \end{pmatrix} \\
&= \frac{\sigma_r^2}{\gamma^2}(J_1^2 + J_2^2 + \cdots + J_s^2) \\
&\stackrel{(23)}{=} \frac{\sigma_r^2}{\sum_{s \in \mathcal{S}_{\text{RV}}} J_s^2}.
\end{aligned} \quad (25)$$